\documentclass[review]{elsarticle}

\usepackage{lineno,hyperref}
\modulolinenumbers[5]

\journal{Artificial Intelligence}

\usepackage{times}
\usepackage{url}
\usepackage{latexsym}
\usepackage{comment}
\usepackage{tikz-dependency}
\usepackage{tikz}
\usepackage{tikz-qtree}
\usepackage[british]{babel}
\usepackage{graphicx}
\usepackage{colortbl}
\usepackage{hhline}
\usepackage{url}
\usepackage{latexsym}
\usepackage{amssymb}
\usepackage{amsmath}
\usepackage{enumerate}
\usepackage{algorithm}
\usepackage{algorithmicx}
\usepackage[noend]{algpseudocode}
\usepackage{booktabs}
\usepackage{proof}
\usepackage{paralist}
\usepackage{tikz-dependency}
\usepackage{tikz} 
\usepackage{graphicx, subfigure}
\usepackage{epstopdf}

\usepackage[latin1]{inputenc}

\newcommand{\eat}[1]{}

\makeatletter
\useshorthands{"}%
\defineshorthand{"-}{\nobreak-\bbl@allowhyphens}
\makeatother

\newcommand{\transname}[1]{\ensuremath{\mathsf{#1}}}

\newcommand{\stacktop}{{\mid}}

\mathchardef\mhyphen="2D 

\newcommand{\sh}{\transname{Shift}}
\newcommand{\re}{\transname{Reduce}}
\newcommand{\reri}{\transname{Reduce\mhyphen Right}}
\newcommand{\rele}{\transname{Reduce\mhyphen Left}}
\newcommand{\reboth}{\transname{Reduce\mhyphen Left/Right}}
\newcommand{\un}{\transname{Reduce\mhyphen Unary}}
\newcommand{\fin}{\transname{Finish}}

\definecolor{red}{rgb}{1,0,0}
\newcommand{\added}[1]{{#1}}

\begin{document}

\begin{frontmatter}

\title{Faster Shift-Reduce Constituent Parsing with a Non-Binary, Bottom-Up Strategy 
\footnote{\textcopyright \ 2019. This is the final peer-reviewed manuscript that was accepted for publication at Artificial Intelligence and made available under the CC-BY-NC-ND 4.0 license \url{http://creativecommons.org/licenses/by-nc-nd/4.0/}. This may not reflect subsequent changes resulting from the publishing process such as editing, formatting, pagination, and other quality control mechanisms. The final journal publication is available at \url{https://doi.org/10.1016/j.artint.2019.07.006}.} }


\author[add1]{Daniel Fern\'{a}ndez-Gonz\'{a}lez\corref{cor1}}
\ead{d.fgonzalez@udc.es}
\cortext[cor1]{Corresponding author.}

\author[add2]{Carlos G\'{o}mez-Rodr\'{i}guez\corref{}}
\ead{carlos.gomez@udc.es}

\address[add1]{Universidade da Coru\~{n}a, 
	FASTPARSE Lab, LyS Research Group, Departamento de Computaci\'{o}n,
	Campus de Elvi\~{n}a, s/n, 15071 A Coru\~{n}a, Spain}

\address[add2]{Universidade da Coru\~{n}a, CITIC. 
	FASTPARSE Lab, LyS Research Group, Departamento de Computaci\'{o}n,
	Campus de Elvi\~{n}a, s/n, 15071 A Coru\~{n}a, Spain}

\begin{abstract}
An increasingly wide range of artificial intelligence applications rely on syntactic information to process and extract meaning from natural language text or speech, with constituent trees being \added{one of the most widely used} syntactic formalisms. To produce these phrase-structure representations from sentences in natural language, shift-reduce constituent parsers have become one of the most efficient approaches. Increasing their accuracy and speed is still one of the main objectives pursued by the research community so that artificial intelligence applications that make use of parsing outputs, such as machine translation or voice assistant services, can improve their performance. With this goal in mind, we propose in this article a novel non-binary shift-reduce algorithm 
for constituent parsing. Our parser follows a classical bottom-up strategy but, unlike others, it straightforwardly creates non-binary branchings with just one $\re$ transition, instead of 
requiring prior binarization or a sequence of binary transitions, allowing its direct application to any language without the need of further resources such as percolation tables. As a result, it uses fewer transitions per sentence than existing transition-based constituent parsers, becoming the fastest such system and, as a consequence, speeding up downstream applications. Using static oracle training and greedy search, the accuracy of this novel approach is on par with state-of-the-art transition-based constituent parsers and outperforms all top-down and bottom-up greedy shift-reduce systems on the Wall Street Journal section from the English Penn Treebank and the Penn Chinese Treebank. Additionally, we develop a dynamic oracle for training the proposed transition-based algorithm, achieving further improvements in both benchmarks and obtaining the best accuracy to date on the Penn Chinese Treebank among greedy shift-reduce parsers.
\end{abstract}


\end{frontmatter}


\section{Introduction}
Natural Language Processing (NLP) aims to transform unrestricted natural language text into a representation that machines can easily handle to provide applications and services widely used nowadays by our society, such as information extraction, machine translation, sentiment analysis or question answering, among others.

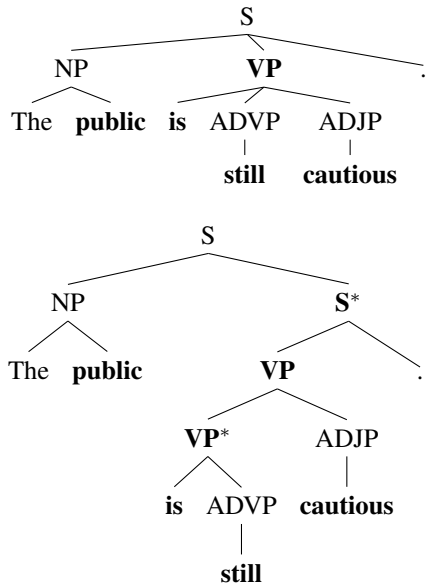
\begin{figure}[t]
\centering
\begin{tikzpicture}[level distance=0.7cm]
\tikzset{frontier/.style={distance from root=76pt}}
\Tree [.S [.{NP} [.The ] [.\textbf{public} ] ]
[.\bf{VP} [.\textbf{is} ] [.ADVP [.\textbf{still} ] ] [.ADJP [.\textbf{cautious} ] ] ] [.. ] ]
\end{tikzpicture}

\vspace{.3cm}
%
\begin{tikzpicture}[level distance=0.9cm]
\tikzset{frontier/.style={distance from root=150pt}}
\Tree [.S [.{NP} [.The ] [.\textbf{public} ] ]
[.\bf{S$^*$} [.\bf{VP} [.\bf{VP$^*$} [.\textbf{is} ] [.ADVP [.\textbf{still} ] ] ] [.ADJP [.\textbf{cautious} ] ] ] [.. ] ] ]
\end{tikzpicture}
\caption{Simplified constituent tree (top) and its binarized equivalent (bottom), taken from English WSJ~\S{22}. Head-child nodes are in bold.}
\label{fig:ctrees}
\end{figure}

\begin{figure}[t]
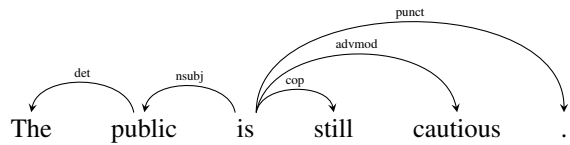

\centering

\begin{dependency}[theme = simple]
\begin{deptext}[column sep=1.8em]
The \& public \& is \& still \& cautious \& . \\
\end{deptext}
\depedge{2}{1}{det}
\depedge{3}{2}{nsubj}
\depedge{3}{4}{cop}
\depedge{3}{5}{advmod}
\depedge[arc angle=90]{3}{6}{punct}
\end{dependency}

\caption{Dependency tree taken from English WSJ~\S{22}.}
\label{fig:dtrees}
\end{figure}

Syntactic parsing is one of these NLP processes that has been in the focus of the research community for the last three decades. This consists in mapping a sentence in natural language into a representation that describes its grammatical structure according to a syntactic formalism. One of the most extended
syntactic formalisms is a \textit{constituent} (or \textit{phrase-structure}) representation \citep{bloomfield33,chomsky56}. Basically, the sentence is decomposed into constituents or phrases and, by creating relationships between words and constituents, a phrase-structure tree is built, like those in Figure~\ref{fig:ctrees}. A simpler alternative is to use \textit{dependency} representations to describe the syntax of a given sentence. This formalism consists of pairs of words linked by binary and asymmetric relations called \textit{dependencies}, denoting which word is the \textit{head} and which is the \textit{dependent}. Figure~\ref{fig:dtrees} depicts an example.

\added{Apart from improving accuracy and coverage, the NLP community is notably interested in increasing parsing speed. 
Syntactic analysis is a crucial task for numerous higher-level artificial intelligence applications, 
such as automatic textual knowledge extraction \citep{Branavan2011}, question answering \citep{CHALI2011}, machine translation \citep{Barone15, Xiao16, Wu17}, automatic  summarization \citep{ABDI2015}, relation extraction \citep{Yu15}, sentiment analysis \citep{SAIF2016,GR2017}, information retrieval \citep{Higashinaka13, FANG2018},  plagiarism detection \citep{K2018}, name entity recognition \citep{Jie17}, among others; and their efficiency is penalized because of the bottleneck resulting from parsers' performance. This is especially dramatic in constituent parsing, whose phrase-based structures are strongly demanded by semantics tasks that involve labeling spans such as semantic role labeling \citep{Gildea02,Swabha18}, coreference resolution \citep{Ng10,Swabha18}, named entity recognition \citep{Finkel09} and also some reading comprehension and question answering tasks \citep{Pranav16}. Constituent parsing has also been widely used in recent works on syntax-based neural network machine translation such as \citep{Eriguchi16, Aharoni17, Li17, Wang18}, that demand an efficient syntactic analysis to accomplish their task.}

Initially, context-free grammar parsers
with  production  rules  derived from data \citep{collins97,charniak96,klein03,petrov07} became popular for generating constituent representations. However, while they provide a good accuracy, they require a significant amount of time to undertake the parsing process, becoming impractical for some downstream applications where the execution time is critical. This led the NLP community to use dependency parsers instead. Despite producing a simpler syntactic representation, their efficiency and speed made them gain popularity, especially linear-time transition-based dependency parsers \citep{nivre03iwpt,nivre04conll, zhang08, Chen2014, Kiperwasser2016, Fernandez18} 
that became the fastest alternative to accurately perform syntactic analysis.

In the past decade, one attempt to speed up constituent parsers was undertaken by using these efficient shift-reduce algorithms to produce more complex structures.
\citet{sagae05} adapted the shift-reduce transition-based framework, 
notably successful
in dependency parsing, to efficiently build constituent 
representations. 

To achieve that, they transformed constituent trees 
making them closer to dependency representations.
In particular, they converted the original 
trees into 
headed, binary constituent trees. In these 
all branchings are 
at most
binary, and nodes are annotated with headedness information, characteristic of dependency syntax \citep{Kahane2015}. 
Figure~\ref{fig:ctrees} shows a constituent tree and its binarized version that describes a syntactic structure similar to that represented by the dependency formalism in Figure~\ref{fig:dtrees} for the same sentence. 

Under this scenario, a transition-based dependency parser such as the \textit{arc-standard} algorithm \citep{Yamada2003,nivre04memorybased} can be easily adapted to produce constituent structures. This 
linear-time shift-reduce algorithm
applies a sequence of transitions that modify the content of two main data structures (a buffer and a stack) and create an arc (in the dependency framework) or connect two child nodes under a non-terminal node (in the constituency framework), building an analysis of the input sentence in a \textit{bottom-up} manner. Further work on this approach  \citep{Zhang2009,Zhu13,Watanabe15,Mi2015,Crabbe2015,Coavoux2016} 
achieved remarkable accuracy and speed, extending its popularity for constituent parsing.

Recently, some researchers have s 
explored non-bottom-up strategies
to build a constituent tree. 
\citep{Dyer2016,Kuncoro2017} proposed a \textit{top-down} transition-based algorithm and, more recently, \citet{Liu2017} developed a
parser that builds the tree in an \textit{in-order} traversal, seeking a compromise between the strengths of top-down and bottom-up approaches, and achieving state-of-the-art accuracy.

\citet{Liu2017} also report that a traditional bottom-up strategy
slightly outperforms
a top-down approach when
implemented under the same framework. This might be 
because,
although a top-down approach adds a lookahead guidance to the 
process, it loses access to the rich 
features from 
partial
trees used in bottom-up parsers.
Thus, it seems that the traditional bottom-up approach 
is still adequate for
phrase structure
parsing and we will show in the following sections how we can notably improve it, increasing not only its accuracy, but also its speed, and make it more appealing for higher-level natural language processing applications.\footnote{An alternative to shift-reduce parsing, which also offers a simpler paradigm for constituent parsing with respect to traditional context-free grammar parsers, is offered by sequence-to-sequence models \citep{Vinyals2015, Ma2017, Liu2018}. However, at the moment these parsers still lag behind shift-reduce parsers in terms of accuracy.}

The remainder of this article is organized as follows: Section~\ref{motivation} presents the main weakness of traditional bottom-up parsers and how we successfully address it. In Section~\ref{preliminaries}, we introduce some notation and concepts about shift-reduce parsers. Section~\ref{approach} describes in detail a novel bottom-up algorithm. 
In Section~\ref{dynamicoracle}, a new method for training our approach is presented and studied. 
Section~\ref{model} presents the neural network architecture used to implement and test the proposed system. In Section~\ref{experiments}, we evaluate in accuracy and speed our approach and compare it against other state-of-the-art shift-reduce systems, and an analysis of the results is provided. Finally, Section~\ref{conclusion} contains a final discussion. 

\section{Motivation and Proposal}
\label{motivation}
Despite being among the most widely-used and efficient constituent parsers, traditional bottom-up shift-reduce algorithms have a notable drawback compared to
other strategies: a binarization as preprocessing is mandatory so that the parser can be applied, and a de-transformation is also required to finally output well-formed phrase-structure trees. 
Apart from the extra time spent, 
the binarization process needs a \textit{percolation table} or a set of \textit{head rules} of the language, which may not always be available. This additional resource consists of a language-specific and handcrafted set of rules that identify which one among the children in a subtree is the head-child (for instance, the head-child of the subtree  $\mathit{NP} \rightarrow \mathit{The}\ \mathit{public}$, present in both trees of Figure~\ref{fig:ctrees}, is $\mathit{public}$) and, as a result, the application of these efficient and accurate systems is restricted to only those languages with a greater amount of resources.

To overcome this, \citep{Cross2016A,Cross2016B} presented transition systems that do not require prior binarization.
However, 
their approach is not
strictly non-binary, as transitions continue affecting only one or two nodes of the tree at a time. For instance, in 
these parsers,
building the ternary branching
$\mathit{VP} \rightarrow \mathit{is}\ \mathit{ADVP}\ \mathit{ADJP}$ in  Figure~\ref{fig:ctrees} takes two 
reduce transitions:
one to connect the two first children under the non-terminal $\mathit{VP}$, and another to add the last child to the subtree. Thus, 
binarization is not needed as pre-processing, but it is applied implicitly during parsing.
Most if not all bottom-up constituent parsing algorithms (not only shift-reduce systems) use binarization explicitly or implicitly, as noted by 
\citet{Gomez2014} for context-free grammar parsers. 

We think that removing any kind of binarization can benefit bottom-up shift-reduce systems' performance; specially in terms of speed, where this kind of algorithms, despite being one of the fastest constituent alternatives, are notably behind dependency parsers. 
In this article, we propose a novel, bottom-up transition-based parser that is purely non-binary: for instance, it can
create the previously mentioned ternary branching at once 
with
a single reduce transition. This not only 
makes transition sequences shorter, achieving faster parsing,
but also 
improves accuracy over
all existing single bottom-up constituent parsers,
explicitly binarized or not. The presented approach also improves over top-down parsers and is on par with state-of-the-art shift-reduce systems on the Wall Street Journal section from the English Penn Treebank \citep{marcus93} and Penn Chinese Treebank (CTB) benchmarks \citep{Xue2005}, being the fastest transition system ever created for constituent parsing.

To further improve the accuracy of our system without harming parsing speed, we also develop a \textit{dynamic oracle} for training it. \citep{Cross2016B,Coavoux2016} have already successfully applied this technique, originally designed for transition-based dependency parsing, to bottom-up phrase-structure parsing. Traditionally, transition-based systems were trained with \textit{static oracles} that follow a gold sequence of transitions to create a model capable of analyzing new sentences at test time. This approach tends to yield parsers that are prone to suffer from error propagation (i.e. errors made in previous parser configurations that are propagated to subsequent states, causing harmful modifications in the  transition sequence). To minimize the effect of error propagation, \citep{goldberg2012dynamic} introduced the idea to train dependency parsers under closer conditions to those found at test time, where mistakes are inevitably made. This can be achieved by training the model with a dynamic oracle with exploration, which allows the parser to go through non-optimal parser configurations during learning time, teaching it how to recover from them and lose the minimum number of gold arcs.

Taking \citep{Coavoux2016} as inspiration, we implement a dynamic oracle for our novel algorithm, achieving notable improvements in accuracy. Unlike \citep{Coavoux2016}, our dynamic oracle is optimal due to the simplicity of our algorithm and the lack of temporary symbols from binarization. 

Experimental results with the dynamic oracle show further improvements in accuracy over the static oracle, and allow us to outperform the state of the art on the CTB supervised setting by 0.7 points, while keeping the same parsing speed.

\section{Preliminaries}
\label{preliminaries}
The basic bottom-up transition system 
of 
\citet{sagae05}, used 
as a base for
many other efficient constituent parsers, analyses a sentence from left to right by reading (moving) words from a buffer to a stack, where partial subtrees are built. 
This is done by
a 
sequence of
$\sh$ (for reading) and $\re$ (for building) transitions that will lead the parser through different states or parser configurations until a terminal one is reached.

More in detail, these parser configurations have the form {$c=\langle {\Sigma} , {i} , {f}, {\gamma} \rangle$, where $\Sigma$ is a \textit{stack} 
of constituents, $i$ is the index of the leftmost word in a list (called \textit{buffer}) of unprocessed words, $f$ is a boolean value that marks if a configuration is terminal or not, and $\gamma$ is the set of constituents 
that have already been built.
Each constituent is represented as a tuple $(X,l,r)$, where $X$ is a non-terminal and $l$ and $r$ are integers defining its span (a word $w_i$ is represented as $(w_i, i, i+1)$). For instance, the non-binary phrase-structure tree on top in Figure~\ref{fig:ctrees} can decomposed as the following set of constituents: \{(S, 0, 6), (NP, 0, 2), (VP, 2, 5), (ADVP, 3, 4), (ADJP, 4, 5)\}.

Note that the information about the set of predicted constituents $\gamma$ and the spans of each constituent is not used by the parser. The transition system can be defined and work without it, and the same applies to the novel non-binary parser that will be introduced below. However, we include it explicitly in configurations because it simplifies the description of our dynamic oracle.

Given an input string $w_0 \cdots w_{n-1}$, the parsing process starts at the initial configuration $c_s(w_0 \ldots w_{n-1}) = \langle [\ ], 0 , \mathit{false}, \{ \} \rangle$ and, after applying a sequence of transitions, it ends in a terminal configuration 
$\langle {\Sigma} , n , \mathit{true}, \gamma \rangle \in C$, where $C$ is the set of possible terminal configurations.

\begin{figure*}

\begin{tabbing}
\hspace{0cm}\=\hspace{2.7cm}\= \kill
\> \sh: 
\> \ \ \ \ \ \ \ $\langle {\Sigma}, {i}, false, \gamma  \rangle
\Rightarrow \langle {\Sigma \stacktop (w_i, i, i+1)} , {i+1}, false, \gamma \cup \{ (w_i, i, i+1) \} \rangle$\\[2mm]
\> \reboth-X:
\>  \ \ \ \ \ \ \ $\langle {\Sigma \stacktop (Y, l, m) \stacktop (Z, m, r)}, i , false, \gamma \rangle
\Rightarrow \langle {\Sigma \stacktop (X, l, r) }, i, false, \gamma \cup \{ (X, l, r) \}  \rangle$ { }\\[2mm]
\> \un-X:
\>   \ \ \ \ \ \ \ $\langle {\Sigma \stacktop  (Y, l, r)}, i , false, \gamma \rangle
\Rightarrow \langle {\Sigma \stacktop (X, l , r) }, i, false, \gamma \cup \{ (X, l, r) \} \rangle$ { }\\[2mm]
\> \fin:
\>   \ \ \ \ \ \ \  $\langle {\Sigma}, n, false, \gamma \rangle
\Rightarrow \langle {\Sigma}, n, true, \gamma \rangle$ { } 
\end{tabbing}

\caption{Transitions of a binary bottom-up constituent parser.}
\label{fig:transitions}
\end{figure*}

Figure~\ref{fig:transitions}
shows
the 
available transitions.
The
$\sh$ transition 
moves the first (leftmost) word in the buffer to the stack;
$\rele$-X and $\reri$-X 
pop the two topmost stack nodes and combine them into a new constituent with the non-terminal X as their parent, 
pushing it onto the stack 
(the head information provided by the direction encoded in each $\re$ transition is used as a feature); 
$\un$ 
pops the top node on the stack, uses it to create a unary subtree with label X, 
which is pushed onto the stack;
and, finally, the $\fin$ transition 
ends the parsing process. Note that every reduction action will add a new constituent to $\gamma$.
Figure~\ref{fig:trans} shows the transition sequence 
that produces
the binary phrase-structure tree in Figure~\ref{fig:ctrees}.

The described transition system is determinized by 
training a classifier to greedily choose which transition should be applied next at each parser configuration. For this purpose, we train the parser to approximate an \textit{oracle}, which chooses optimal transitions with respect to gold parse trees.
This oracle can be static or dynamic, depending on the strategy used for training the parser. A static oracle trains the parser on only gold parser configurations, while the dynamic one can train the parser in any possible configuration, allowing the exploration of non-optimal parser states.

\begin{figure*}
\begin{center}
\vspace*{13pt}
\begin{tabular}{lccc}
\hline\noalign{\smallskip}
Transition & Stack & Buffer & Added subtree \\
\noalign{\smallskip}\hline\noalign{\smallskip}
 & [ ] & [ The, ... , \textbf{.} ] &  \\
\textsc{$\sh$} & [ The ] & [ public, ... , \textbf{.} ] &  \\
\textsc{$\sh$} & [ The, public ] & [ is, ... , \textbf{.} ] &  \\
\textsc{$\rele$-NP} & [ NP ]  & [ is, ... , \textbf{.} ] & NP$\rightarrow$The public \\
\textsc{$\sh$} & [ NP, is ]  & [ still, ... , \textbf{.} ] &  \\
\textsc{$\sh$} & [ NP, is, still ]  & [ cautious , \textbf{.} ] &  \\
\textsc{$\un$-ADVP} & [ NP, is, ADVP ]  & [ cautious , \textbf{.} ] & ADVP$\rightarrow$ still \\
\textsc{$\reri$-VP$^*$} & [ NP, VP$^*$ ]  & [ cautious , \textbf{.} ] & VP$^*$$\rightarrow$is ADVP \\
\textsc{$\sh$} & [ NP, VP$^*$, cautious ]  & [ \textbf{.} ] &  \\
\textsc{$\un$-ADJP} & [ NP, VP$^*$, ADJP ]  & [ \textbf{.} ] & ADJP$\rightarrow$ cautious \\
\textsc{$\reri$-VP} & [ NP, VP ]  & [ \textbf{.} ] & VP$\rightarrow$VP$^*$ ADJP  \\
\textsc{$\sh$} & [ NP, VP, \textbf{.} ]  & [ ] &  \\
\textsc{$\reri$-S$^*$} & [ NP, S$^*$ ]  & [ ] & S$^*$$\rightarrow$VP \textbf{.} \\
\textsc{$\rele$-S} & [ S ]  & [ ] & S$\rightarrow$NP S$^*$ \\
\textsc{$\fin$} & [ ]  & [ ] &  \\
\noalign{\smallskip}\hline
\end{tabular}
\caption{Transition sequence for producing the binary constituent tree in Figure~\ref{fig:ctrees} using
a traditional binary bottom-up parser.} \label{fig:trans}       
\vspace*{13pt}
\end{center}
\end{figure*}

\section{Novel Bottom-up Transition System}
\label{approach}
We propose a novel transition system for bottom-up constituent parsing that 
builds more-than-binary branchings
at once by applying a single $\re$ transition. We keep the parser configuration form defined by Sagae and Lavie \citep{sagae05}, but modify the transition set. In particular, we design a new non-binary $\re$ transition that can create a subtree rooted at a non-terminal X and with the $k$ topmost stack nodes as its children, with $k$ ranging from $1$ (to produce unary branchings) to the sentence length $n$. In that way, the ternary branch $\mathit{VP} \rightarrow \mathit{is}\ \mathit{ADVP}\ \mathit{ADJP}$, used previously as an example, could be created by just applying a $\re$-VP\#3 transition, which will build a new constituent VP with three children at once. 

The proposed transition set is formally described in Figure~\ref{fig:transitions2}.
Note that no specific transition is required for producing unary branching, 
as the
novel $\re$ transition can handle any kind of phrase-structure trees.

This approach does not need any kind of previous binarization 
(contrary to the one described in
Figure~\ref{fig:transitions}) and, therefore, it lacks headedness information provided by binarized trees. In addition, since constituents of any kind can be reduced with just one transition, the parsing of a given sentence is done by consuming the shortest sequence of transitions among 
known transition systems
for constituent parsing. Figure~\ref{fig:trans2} shows the transition sequence necessary to produce the non-binary structure in Figure~\ref{fig:ctrees}. 
In that simple example,
we use 12 transitions,
while the binary bottom-up system consumes 14 to parse the same sentence.
Apart from being slower, said binary bottom-up parser 
needs
to apply 
an unbinarization process
to the parser output to produce a final non-binary 
phrase structure 
tree. The top-down \citep{Dyer2016} and the in-order \citep{Liu2017} algorithms 
need even more transitions to produce the 
tree in Figure~\ref{fig:ctrees}: 16 and 17, respectively.

Moreover, this novel non-binary $\re$ transition 
naturally lends itself to handling
a non-terminal in a different way depending on its arity,
enlarging the non-terminal dictionary. For instance, our system 
can distinguish 
between a verbal phrase with two children (VP\#2) and 
one
with three children (VP\#3) and treat them differently. 
This is a way of encoding some information
about grammatical subcategorization frames in the non-binary $\re$ transition, which can be useful for the parser to learn in what circumstances it should create a VP\#2 with only two elements (for instance, the verb and its direct object) 
or three (for example, if that particular verb 
is ditransitive, and subcategorizes for both direct and indirect objects).
To achieve that, we use different vector representations for non-terminals VP\#2 and VP\#3.

\added{It is worth mentioning that this straightforward non-binary algorithm could not have been successfully applied on pre-deep-learning techniques, mainly due to the considerable amount of labels that the model has to deal with.}

\begin{figure*}
\begin{tabbing}
\hspace{0cm}\=\hspace{2.3cm}\= \kill
\> \sh:
\>   \ \ \ \ \ \ \   $\langle {\Sigma}, {i}, false, \gamma \rangle
\Rightarrow \langle {\Sigma \stacktop (w_i, i, i+1)} , {i+1}, false, \gamma \cup \{ (w_i, i, i+1) \} \rangle$\\[2mm]
\> \re-X\#$k$:
\>   \ \ \ \ \ \ \   $\langle {\Sigma \stacktop (Y_{1},m_0, m_1) \stacktop ... \stacktop (Y_k, m_{k-1}, m_k)}, i , false, \gamma \rangle \Rightarrow$\\[2mm]
\> \hspace{6.2cm}$\langle {\Sigma \stacktop (X, m_0, m_k), i, false, \gamma \cup \{ (X, m_0, m_k) \} } \rangle$ { }\\[2mm]
\> \fin:
\>   \ \ \ \ \ \ \   $\langle {\Sigma}, n, false, \gamma \rangle
\Rightarrow \langle {\Sigma}, n, true, \gamma \rangle$ { }
\end{tabbing}

\caption{Transitions of the novel non-binary bottom-up constituent parser.}
\label{fig:transitions2}
\end{figure*}

\begin{figure*}
\begin{center}
\vspace*{13pt}
\begin{tabular}{lccc}
\hline\noalign{\smallskip}
Transition & Stack & Buffer & Added subtree \\
\noalign{\smallskip}\hline\noalign{\smallskip}
 & [ ] & [ The, ... , \textbf{.} ] &  \\
\textsc{$\sh$} & [ The ] & [ public, ... , \textbf{.} ] &  \\
\textsc{$\sh$} & [ The, public ] & [ is, ... , \textbf{.} ] &  \\
\textsc{$\re$-NP\#2} & [ NP ]  & [ is, ... , \textbf{.} ] & NP$\rightarrow$The public \\
\textsc{$\sh$} & [ NP, is ]  & [ still, ... , \textbf{.} ] &  \\
\textsc{$\sh$} & [ NP, is, still ]  & [ cautious , \textbf{.} ] &  \\
\textsc{$\re$-ADVP\#1} & [ NP, is, ADVP ]  & [ cautious , \textbf{.} ] & ADVP$\rightarrow$ still \\
\textsc{$\sh$} & [ NP, is, ADVP, cautious ]  & [ \textbf{.} ] &  \\
\textsc{$\re$-ADJP\#1} & [ NP, is, ADVP, ADJP ]  & [ \textbf{.} ] & ADJP$\rightarrow$ cautious \\
\textsc{$\re$-VP\#3} & [ NP, VP ]  & [ \textbf{.} ] & VP$\rightarrow$is ADVP ADJP  \\
\textsc{$\sh$} & [ NP, VP, \textbf{.} ]  & [ ] &  \\
\textsc{$\re$-S\#3} & [ S ]  & [ ] & S$\rightarrow$NP VP \textbf{.} \\
\textsc{$\fin$} & [ ]  & [ ] &  \\
\noalign{\smallskip}\hline
\end{tabular}
\caption{Transition sequence for producing the constituent tree in Figure~\ref{fig:ctrees} using
the proposed non-binary bottom-up parser.} \label{fig:trans2}   
\vspace*{13pt}
\end{center}
\end{figure*}

\subsection{\added{Time Complexity}}
\added{Existing shift-reduce transition systems, such as \citep{sagae05,Dyer2016,Liu2017}, tend to have a linear running time complexity with respect to the length of the input sentence; however, the non-binary algorithm presented here has a quadratic theoretical complexity in the worst case.} 

\added{To reach this result, we observe that the non-binary transition-based algorithm can parse any sentence with length $n$ by applying $n$ $\sh$ transitions to read every word from the input, plus $|N|$ $\re$ transitions to build every subtree (with $N$ being the set of non-terminal nodes in the output tree), plus one $\fin$ transition that ends the process. Therefore, the number of transitions required for analyzing a sentence of length $n$ is exactly $n+|N|+1$.}

\added{To write $|N|$ as a function of $n$, we assume that the length of chains of unary nodes in constituent trees are bounded by a constant $k$ (without this assumption, constituent trees can be of arbitrary size even if $n$ is bounded, and thus complexity of this or any other constituent parser cannot be bounded). Under this assumption, $|N|$ is at most $k(n-1)$ (the worst case is a complete binary tree, with $n-1$ nonterminal nodes, which are then expanded with the maximum allowable unary chains) and thus the total number of transitions in the worst case is $n+k(n-1)+1$, which is $O(n)$.}

\added{In addition, let $c$ be the maximum number of children that a non-terminal can have. Then, up to $c$ $\re$ transitions with different arity might be evaluated at each step in the worst case. Since $c$ can never be larger than $n$, in the worst case we have $O(n)$ steps where $O(n)$ transitions need to be evaluated, and thus the worst-case complexity is $O(n^2)$.}

\added{In spite of that, in section~\ref{sec:run}, we show empirically that the non-binary algorithm behaves as a linear parser in practice, since in practice $c$ is much smaller than $n$ and behaves like a constant: the number of possible children of a given non-terminal (and, therefore, the number of possible parametrized $\re$ transitions that will be evaluated) tends to stay low throughout the training dataset.}

\section{Dynamic Oracle}
\label{dynamicoracle}
Dynamic oracles have been thoroughly studied for a large range of existing transition systems in dependency parsing. However, only a few papers show some progress in constituent parsing \citep{Cross2016B,Coavoux2016}.

Broadly,
implementing a dynamic oracle \citep{goldberg2012dynamic} 
consists of
defining a \textit{loss function} 
on configurations, measuring the distance from the best tree they can produce to the gold parse.
In that way, we can compute the cost of the new configurations resulting from applying each permissible transition, thus obtaining which transitions will lead the parser to configurations where the minimum number of errors are made.

More formally, given a parser configuration $c$ and a gold tree $t_G$, a loss function $\ell(c)$ is usually implemented as the minimum Hamming loss between $t$ and $t_G$, ($\mathcal{L}(t,t_G)$), where $t$ is the already-built tree of a configuration $c'$ reachable from $c$ (written as $c \rightsquigarrow t$). The Hamming loss is computed in dependency parsing as the amount of nodes in $t$ with a different head in $t_G$. Therefore, the loss function 
defined as:
\[ \ell(c) = \min_{t | c \rightsquigarrow t} \mathcal{L}(t,t_G) \] 
\noindent will compute the cost of a configuration as the Hamming loss of the reachable tree $t$ that differs the minimum from $t_G$.

A correct dynamic oracle will return the set of transitions $\tau$ that do not increase the overall loss (i.e., $\ell( \tau(c) ) - \ell(c) = 0 $) and, thus, will lead the system through optimal configurations, minimizing the Hamming loss with respect to $t_G$.

This same idea can be applied in constituent parsing, as done by \citep{Coavoux2016},
if we redefine the Hamming loss to work with
constituents instead of arcs. In this case, $\mathcal{L}(t,t_G)$ 
is defined as the
size of the symmetric difference between the constituents in the trees $t$ and $t_G$.

To build a correct oracle, we need to find an easily-computable expression for this minimum Hamming loss. 
For this purpose, we will study constituent reachability. Namely, we will show that this parser has the constituent decomposability property \citep{Coavoux2016}, analogous to the arc-decomposability property of some dependency parsers \citep{goldberg2013training}. This property implies that $|t_G \setminus t|$ can be obtained simply by counting the individually unreachable constituents from configuration $c$, greatly facilitating the definition of the loss function as the other term of the symmetric difference $(|t \setminus t_G|)$ is easy to minimize.

\subsection{Constituent Reachability}
To reason about constituent reachability and decomposability, we will represent phrase structure trees as a set of constituents.
As we have seen above, the non-binary phrase-structure tree in Figure~\ref{fig:ctrees} can be decomposed as this set of constituents: \{(NP, 0, 2), (VP, 2, 5), (ADVP, 3, 4), (ADJP, 4, 5) and (S, 0, 6)\}.  

Let ${\gamma_G}$ be the set of gold constituents for our current input. We say that a gold constituent $(X, l, r) \in {\gamma_G}$ is reachable from a configuration $c=\langle {\Sigma} , {j} , \mathit{false}, {\gamma_c}\rangle$ with $\Sigma = [(Y_p, i_p, i_{p-1})\cdots(Y_{2}, i_{2}, i_{1})|(Y_1, i_{1}, j)]$, and it is included in the set of \textit{individually reachable constituents} $\mathcal{R}(c,{\gamma}_G)$, iff it satisfies one of the following conditions:}
\begin{itemize}
\item $(X, l, r) \in \gamma_c$ (i.e. it has already been created and, therefore, it is reachable by definition).
\item $j \leq l < r$ (i.e. it is still in the buffer and can be still created after shifting more words). 
\item $l \in \{i_k \mid 1 \le k \le p\} \wedge j \leq r$
(i.e. its span is completely or partially in the stack, sharing left endpoint with the $k$th topmost stack constituent, so it can still be built in the future by a transition reducing the stack up to and including that constituent).
\end{itemize}
The set of \textit{individually unreachable constituents} $\mathcal{U}(c,{\gamma}_G)$ with respect to the set of gold constituents ${\gamma}_G$ can be easily computed as $ {\gamma}_G \setminus \mathcal{R}(c,{\gamma}_G)$ and will contain the gold constituents that can no longer be built, be it because we have read past their span or because we have created constituents that would overlap with them.

\subsection{Loss function}
In dependency parsing it is not necessary to count false positives as separate errors from false negatives, as a node attached to the wrong head (false positive) is directly tied to some gold arc being missed (false negative) due to the single-head constraint. 
In phrase-structure parsing, however, a parser can incur extra false positives by creating erroneous extra constituents, harming precision. For this reason, just like the standard F-score metric penalizes false positives, the loss function must also do so. Hence, as mentioned earlier, we base our loss function in the symmetric difference between reachable and gold constituents.

Thus, our loss function is
\[ \ell(c) = \min_{\gamma | c \rightsquigarrow \gamma} \mathcal{L}(\gamma,\gamma_G) = |{\mathcal{U}(c,{\gamma}_G)}| + |{\gamma_c} \setminus {\gamma}_G| \\\]

\noindent where the first term ($|\mathcal{U}(c,{\gamma}_G)|$) penalizes false negatives (gold constituents that we are guaranteed to miss, as they are unreachable), and the second term ($|{\gamma_c} \setminus {\gamma}_G|$) penalizes false positives (erroneous constituents that have been built).

It is worth mentioning that we cannot directly apply the cost formulation defined by Coavoux and Crabb\'e \citep{Coavoux2016} as they rely on the fixed number of constituents 
in a binary phrase-structure tree to implicitly penalize the prediction of non-gold binary constituents (false positives), and thus only need to explicitly penalize wrong unaries with a dedicated term. 
In our case, the second term of our loss function is used to penalize the creation of any kind of non-gold constituents, unary or not.

In addition, unlike \citep{Coavoux2016}, our expression 
provides the exact loss
due to the lack of dummy temporary symbols required by the binarization (they use a traditional binary bottom-up transition system).

\subsection{Optimality} 

We now prove the correctness (or optimality) of our oracle, i.e., that the expression of $\ell(c)$ defined above indeed provides the minimum possible Hamming loss to the gold tree among the trees reachable from a configuration $c$:
\[ \min_{\gamma | c \rightsquigarrow \gamma} \mathcal{L}(\gamma,\gamma_G) = |{\mathcal{U}(c,{\gamma}_G)}| + |{\gamma_c} \setminus {\gamma}_G| \\\]
First, we show that the algorithm is constituent-decomposable, i.e., that if each of a set of $m$ constituents is individually reachable from a configuration $c$, then the whole set also is. We prove this by induction on $m$. The case for $m=1$ is trivial. Let us now suppose that constituent-decomposability holds for sets of size up to $m$, and show that it also holds for an arbitrary set $T$ of $m+1$ tree-compatible constituents.

Let $(X,l,r)$ be one of the constituents of $T$ such that $r = \min \{ r' \mid (X',l',r') \in T \}$ and $l = \max \{ l' \mid (X',l',r) \in T\}$.
Let $T' = T \setminus (X,l,r)$. $T'$ has $m$ constituents, so by induction hypothesis, it is a reachable set from configuration $c$.

By hypothesis, $(X,l,r)$ is individually reachable, so at least one of the three conditions in the definition of the individual reachability must hold.

If the first condition holds, then $(X,l,r)$ has already been created in $c$. Thus, any transition sequence that builds $T'$ (which is a reachable set) starting from $c$ will also include $(X,l,r)$, so $T = T' \cup \{(X,l,r)\}$ is reachable from $c$.

If the second condition holds, then $j \le l < r$ and we can create $(X,l,r)$ with $r-j$ $\sh$ transitions followed by one \re-X\#$(r-l)$ transition. Constituents of $T'$ are still individually reachable in the resulting configuration, as they either have right index at least $r$ and left index at most $l$ (so that the third reachability condition) or right index greater than $r$ and left index at least $r$ (so they meet the second).
Hence, reasoning as in the previous case, $T$ is reachable from $c$.

Finally, if the third condition holds, then $l$ is the left endpoint of the $k$th topmost stack constituent and $r \ge j$. Then, we can create $(X,l,r)$ with $r-j$ $\sh$ transitions followed by one \re-X\#$(r-j+k)$ transition. By the same reasoning as in the previous case, $T$ is reachable from $c$.

With this we have shown the induction step, and thus constituent decomposability. This implies that, given a configuration $c$, there is always some transition sequence that builds all the individually reachable constituents from $c$, missing only the individually unreachable ones, i.e., $\min_{\gamma | c \rightsquigarrow \gamma} |\gamma_G \setminus \gamma| = |{\mathcal{U}(c,{\gamma}_G)}|$.

To conclude the correctness proof, we note that the set $(\gamma \setminus \gamma_G)$ (false positives) will always contain $(\gamma_c \setminus \gamma_G)$ for $c \rightsquigarrow \gamma$ 
(as the algorithm is monotonic and never deletes constituents); and there exists at least one transition sequence from $c$ that generates exactly the tree $\mathcal{R}(c,{\gamma}_G) \cup (\gamma_c \setminus \gamma_G)$ (as the algorithm has no situations such that creation of a constituent is needed as a precondition for another). This tree has loss $|{\mathcal{U}(c,{\gamma}_G)}| + |{\gamma_c} \setminus {\gamma}_G|$, which is thus the minimum loss.

It is worth noting that the correctness of our oracle contrasts with the case of \citep{Coavoux2016}, whose oracle is only correct under the ideal case where there are no temporary symbols in the grammar, so that they have to resort to a heuristic for the general case.

\section{Neural Model}
\label{model}
We implement our system 
with greedy decoding
under the neural transition-based
framework developed 
by \citet{Dyer2016} and reused by \citet{Liu2017}. \added{This provides a state-of-the-art neural network architecture that proved to be one of the best options to date for implementing transition-based constituent parsers. Another possible alternative could be a BiLSTM-based architecture \citep{Kiperwasser2016,Cross2016A}, which has also achieved good accuracy in constituent parsing, as reported by \citep{Cross2016B}}.  

In particular, the framework introduced by \citet{Dyer2016} follows a stack-LSTM \citep{Dyer2015} approach to represent the stack and the buffer, where each element is the result of a compositional representation  of  partial  trees. In addition, a vanilla LSTM \citep{Hochreiter97} is in charge of representing the action history. 

This architecture is trained for greedily and sequentially making local decisions. Given a sentence $w_0, w_1, ... , w_{n-1}$ with $n$ words, and with $w_i$ being the $i$th word, for the $k$th parsing state [$c_j, ... , c_1, c_0$, $i$, $\mathit{false}$] the distributional probability of the current action $p$ is computed as:
\begin{multline*}
{p = \mathit{softmax}(W\ E_{state} + b)},
\end{multline*}

\noindent where $W$ and $b$ are model parameters, and $E_{state}$ is the embedding that represents the current parsing state and is obtained by the concatenation and posterior transformation of the representations of elements that compound the transition system as follows:
\begin{multline*}
{E_{state} = \mathit{relu}(W_s[E_{stack}, E_{buffer}, E_{history}] + b_s)}
\end{multline*}

\begin{figure}
\begin{center}
\includegraphics[width=\columnwidth]{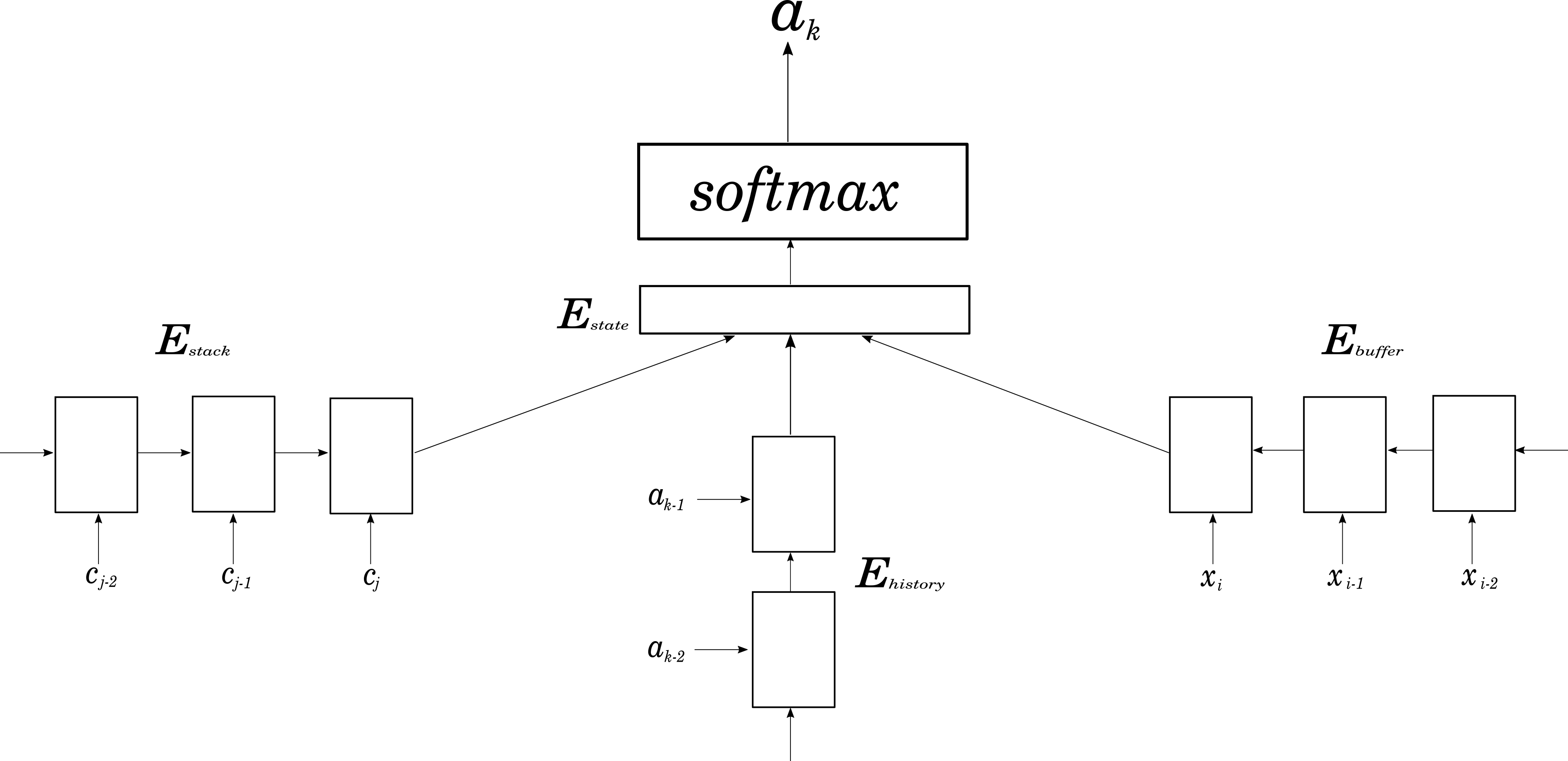}
\end{center}
\caption{Neural architecture for greedy shift-reduce constituent parsing.}
\label{fig:model}
\end{figure}

\noindent Note that $W_s$ and $b_s$ are model parameters and $E_{stack}$ is the vector representation obtained from the stack-LSTM that stores the constituents currently in the stack:
\begin{multline*}
{E_{stack} = \mathit{stack-LSTM}[c_0, c_1, ... , c_j]},
\end{multline*}
$E_{buffer}$ is the vector representation of the current buffer as:
\begin{multline*}
{E_{buffer} = \mathit{stack-LSTM}[x_i, x_{i+1}, ... , x_{n-1}]},
\end{multline*}
where $x_i$ is a word representation, and finally, the representation of action history is computed as:
\begin{multline*}
{E_{history} = \mathit{LSTM}[a_{k-1}, a_{k-2}, ... , a_0]}
\end{multline*}
\noindent where $a_t$ is the vector representation of action in $t$th parsing state. The detailed neural network architecture is depicted in Figure~\ref{fig:model}.

Following \citep{Dyer2016} and \citep{Liu2017}, the set of parameters $\theta$ in the model are trained to minimize a cross-entropy loss objective with an $l_2$ regularization term defined by: 
\begin{multline*}
{L(\theta) = - \sum_{i}\sum_{j}log\ p_{ij} + \frac{\lambda}{2} ||\theta||^2} 
\end{multline*}

\noindent where $p_{ij}$ is the probability of the $j$th action in the $i$th training example provided by the model and $\lambda$ is the regularization hyper-parameter. Finally, stochastic gradient descent is used for optimizing $\theta$.

\subsection{Word Representation Strategy}
We follow the same word representation strategy as described by \citet{Dyer2015}. More in detail, we combine three different embeddings: pretrained word embeddings ($e^{*}_{w_i}$), randomly initialized word embeddings ($e_{w_i}$) and 
randomly initialized POS tag embeddings ($e_{p_i}$). Those embeddings randomly initialized are fine-tuned during the training. After being concatenated the three embeddings, an affine transformation and a posterior non-linear function ReLu is used to derive the final vector representation $x_i$:
\begin{multline*}
{x_i = relu(W_e[e^{*}_{w_i}, e_{w_i}, e_{p_i}] + b_e)}
\end{multline*}
\noindent where $W_e$ and $b_e$ are model parameters, and $w_i$ and $p_i$ represent the form and the POS tag of the $i$th input word.

\subsection{Syntactic Compositional Function}
\begin{figure}
\begin{center}
\includegraphics[width=\columnwidth]{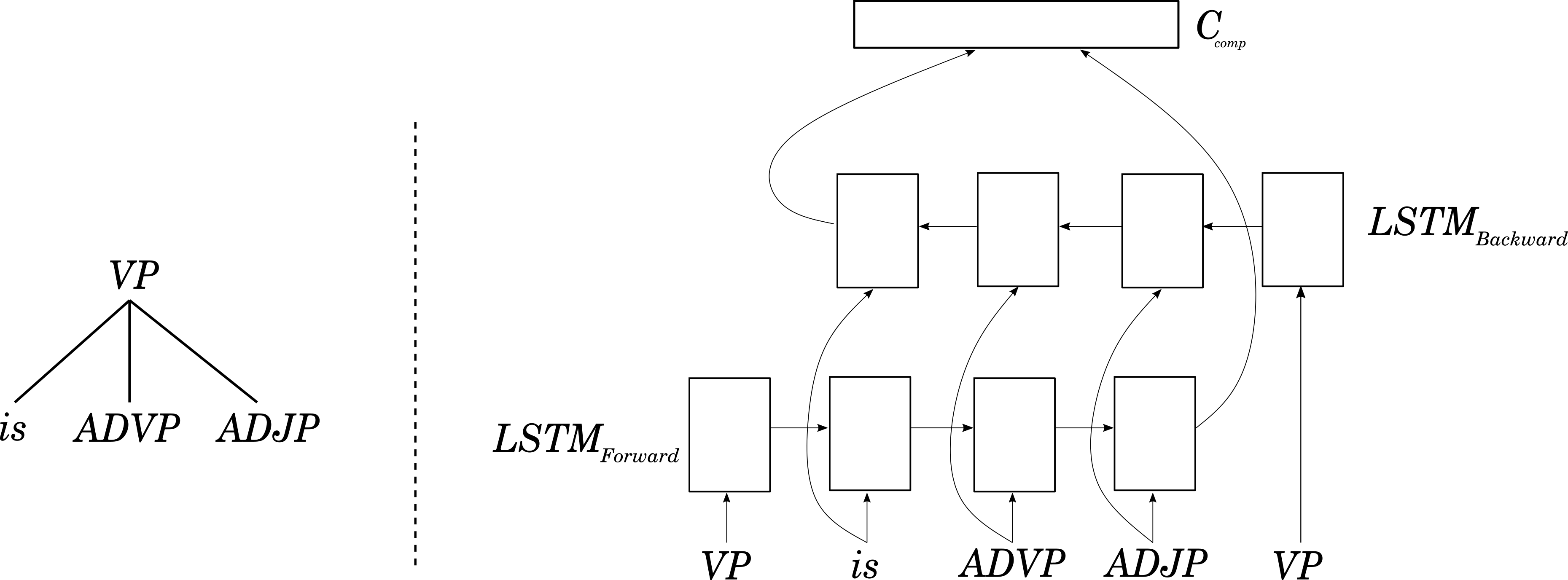}
\end{center}
\caption{Non-binary syntactic compositional function that reduces a subtree from Figure~\ref{fig:ctrees} into a single vector representation $c_{comp}$ by applying a BiLSTM over the tree elements' embeddings.}
\label{fig:comp}
\end{figure}

Following \citet{Dyer2016}, when a $\re$
transition is applied,  the parser
pops  a  sequence  of  completed  subtrees  and/or  words (represented as vector embeddings) from
the stack and makes them children of the selected non-terminal,  building a new constituent.  In order to obtain a vector representation for this new
subtree,  we  use  a syntactic  compositional  function  based  on
bidirectional LSTMs (BiLSTMs) \citep{Graves05}. 
For the new non-binary bottom-up parser we use the method originally applied in \citep{Dyer2016} for the top-down algorithm, and adopted by \citep{Liu2017} for the in-order parser. More in detail, both LSTMs that compound the BiLSTM read first the selected non-terminal node and, then, the elements from the stack involved in the $\re$ transition are introduced in forward direction in one of the LSTMs and, in reverse order, in the other, as shown graphically in Figure~\ref{fig:comp}. After that, the final states of the forward and backward LSTMs are concatenated, and an affine transformation and a ReLu non-linearity are applied to finally compute the composition representation $c_{comp}$:
\begin{multline*}
{c_{comp} = relu(W_c[LSTM_{fwd}[e_{nt}, c_0, ... , c_m]; LSTM_{bwd}[e_{nt}, c_m, ... , c_0]] + b_c )}
\end{multline*}
\noindent where $W_c$ and $b_c$ are model parameters, $e_{nt}$ is the vector representation of the non-terminal on top of the new constituent, and $c_i$, $i \in [0,m]$ is the vector representation of the $i$th child node, which might be the vector embedding of a completed subtree or a word.

\begin{figure}
\begin{center}
\includegraphics[width=\columnwidth]{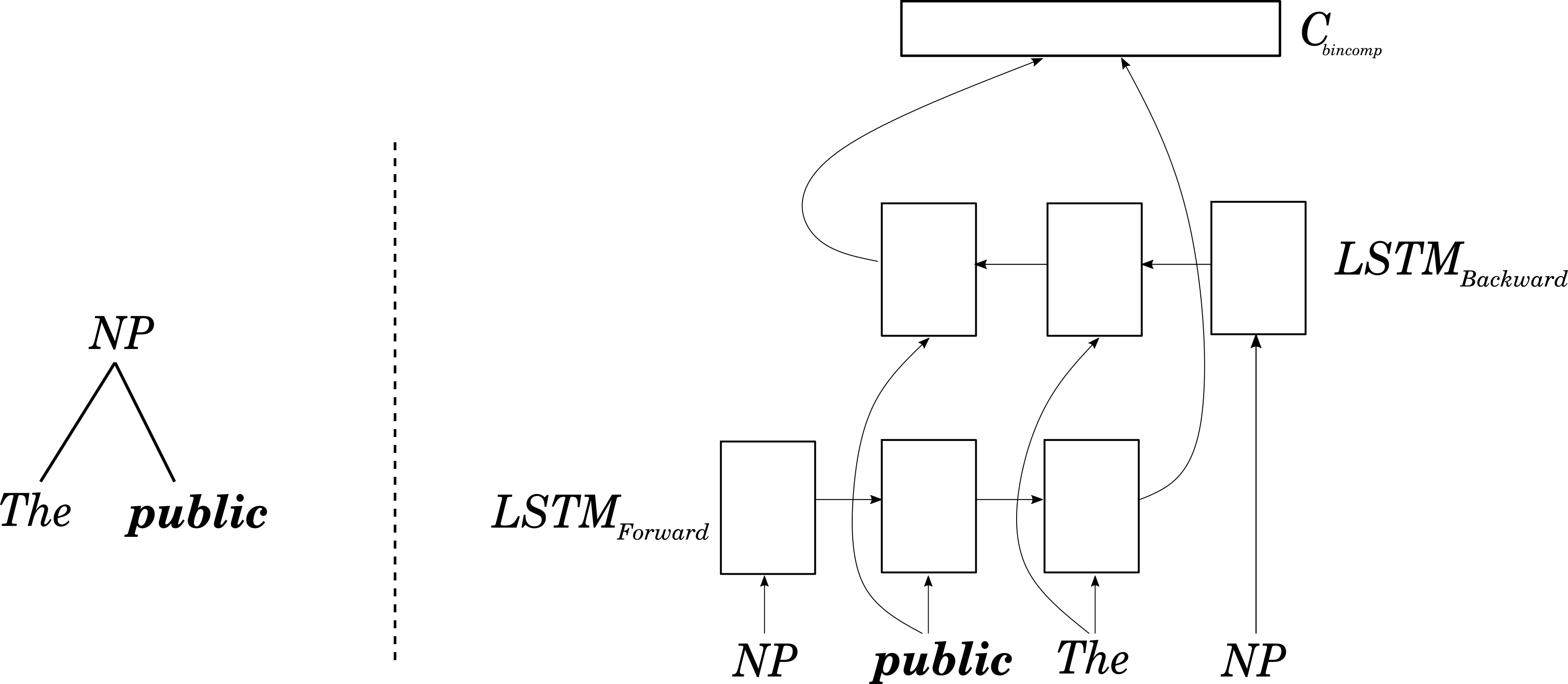}
\end{center}
\caption{Binary syntactic compositional function to reduce a subtree from Figure~\ref{fig:ctrees} into a single vector representation $c_{bincomp}$. Note that the head-child marked in bold is read first by the BiLSTM.}
\label{fig:comp2}
\end{figure}

This differs from the compositional function used for the binary bottom-up parser implemented by \citep{Liu2017} under the same framework. More in detail, the composition representation of the new constituent $c_{bincomp}$ for the binary case is computed as:
\begin{multline*}
{c_{bincomp} = relu(W_c[LSTM_{fwd}[e_{nt}, c_h, c_c]; LSTM_{bwd}[e_{nt}, c_c, c_h]] + b_c)}
\end{multline*}
\noindent where $c_h$ and $c_c$ denote the representations of the constituents acting as head-child and as a regular child, respectively. Unlike in the non-binary case, in this compositional function we need to use a percolation table to explicitly mark which constituent is the head-child node that must be first in the BiLSTM, adding additional information to the training. Figure~\ref{fig:comp2} depicts graphically how the compositional function works in the binary scenario.

\added{Please note that the proposed transition system and the framework used are orthogonal to 
approaches like beam search, re-ranking or semi-supervision that might easily improve the final accuracy; however, all these techniques would penalize parsing speed.}

\section{Experiments}
\label{experiments}
\subsection{Data and Settings}
We conduct our experiments on 
two widely-used benchmarks
for evaluating constituent parsers: the Wall Street Journal (WSJ) sections of
the English Penn Treebank \citep{marcus93} (Sections  2-21 are used as training data, Section 22 for development and Section 23 for testing) and version 5.1 of the Penn Chinese Treebank (CTB) \citep{Xue2005} (articles 001- 270 and 440-1151 are taken for training, articles 301-325 for system development, and articles 271-300 for final testing). 

We use the same POS tags as \citep{Dyer2016} and \citep{Liu2017} automatically predicted by the Stanford tagger \citep{Toutanova2003} for English and by Marmot \citep{Marmot} for Chinese. The same pre-trained word embeddings  are also adopted: those generated on the AFP portion of English Gigaword corpus (version 5) and those extracted from the complete Chinese Gigaword corpus (version 2).

Finally, we use exactly the same hyper-parameter values as \citep{Dyer2016} and \citep{Liu2017} without further optimization. We summarize them in Table~\ref{tab:hyperparameters}.

\begin{table}
\begin{center}
\centering
\begin{tabular}{@{\hskip 0.5pt}lc@{\hskip 0.5pt}}
Hyper-parameter & Value\\
\hline
Initialized learning rate & 0.1 \\
Learning rate decay & 0.05 \\
$\lambda$ &  10$^{-6}$ \\
Dropout rate & 0.2 \\
Number LSTM layers & 2 \\
Stack-LTSM input dimension & 128 \\
Stack-LTSM hidden dimension & 128 \\
Word embedding dimension & 32 \\
POS tag embedding dimension & 12 \\
English pretrained word embedding dimension & 100 \\
Chinese pretrained word embedding dimension & 80 \\
Action embedding dimension & 16 \\
\hline
\multicolumn{1}{c}{}\\
\end{tabular}
\caption{Hyper-parameters used for all experiments.
}
\label{tab:hyperparameters}
\end{center}
\end{table}

\subsection{Error exploration}

Success of dynamic oracles relies on the use of a good exploration strategy. Some recent dependency parsing approaches follow 
\citet{Kiperwasser2016}, 
who undertake aggressive exploration to increase the impact of error exploration and avoid the early overfitting of the training data by neural networks. 
In particular, their implementation chooses
a non-optimal action when either of these two conditions are satisfied: (1) its score is lower than the score of the optimal one by at least $\alpha$ (i.e. we do not have a strong optimal action and go for a non-optimal one, called aggressive exploration criterion parametrized by a constant margin $\alpha$) or (2) its score is higher, with probability $\beta$ (i.e. the non-optimal transition is the highest-scoring one and we follow it, but with a low probability, and we call this regular exploration criterion parametrized by a probability $\beta$).

We ran a few experiments on the development set to check which strategies and parameters were more suitable for our dynamic oracle. 
We started by considering
the setting 
used by \citep{Kiperwasser2016} (aggressive
exploration with margin 1.0 together with regular exploration with
probability 0.1)
and, as an alternative, we studied 
less aggressive approaches
that use only regular exploration with a small range of values of $\beta$.
As shown in Table~\ref{tab:error}, aggressive exploration yields the highest F-score on the CTB, while a more conservative strategy (regular exploration with probability 0.2) achieves better results on the WSJ.

\begin{table}
\begin{center}
\begin{tabular}{lcc}
Strategy & WSJ & CTB \\
\hline
\small{aggr-1.0 $\vee$ reg-0.1} & 91.83 & \textbf{89.86}\\
\small{reg-0.1} & 91.81 & 89.70 \\
\small{reg-0.2} & \textbf{91.88} & 89.47 \\
\small{reg-0.3} & 91.82 & 89.69 \\
\hline
\multicolumn{1}{c}{}\\
\end{tabular}
\caption{F-score comparison of different error-exploration strategies on WSJ~\S22 and CTB~\S301-325. Note that ``aggr-$\alpha$'' stands for aggressive-exploration  and,``reg-$\beta$'', for regular-exploration.
}
\label{tab:error}
\end{center}
\end{table}

\begin{table}
\begin{center}
\centering
\begin{tabular}{@{\hskip 0.5pt}lcccc@{\hskip 0.5pt}}
Parser & Bin & Type & Strat & F1 \\
\hline
\citet{Cross2016A} & n & gs & bu & 90.0  \\
\citet{Cross2016B} & n & gs & bu & 91.0  \\
\citet{Cross2016B} & n & gd & bu & 91.3  \\
\citet{Liu2017}  & y & gs & bu & 91.3  \\
\textbf{This work} & n & gs &  bu & \textbf{91.5}\\ 
\textbf{This work} & n & gd &  bu & \textbf{91.7}\\
\hline
\citet{Zhu13} & y & b & bu & 90.4  \\
\citet{Watanabe15} & y & b & bu & 90.7  \\
\citet{Liu2017B} & y & b & bu & 91.7    \\
\hline
\citet{Dyer2016} & n & gs & td & 91.2    \\
\citet{Liu2017}  & n & gs & in & \textbf{91.8}   \\
\hline
\multicolumn{1}{c}{}\\
\end{tabular}

\caption{Accuracy comparison of state-of-the-art shift-reduce constituent parsers on WSJ~\S{23}. The ``Bin'' column marks if prior explicit binarization is required ($y$es/$n$o). The ``Type'' column shows the type of parser: \emph{gs} is a greedy parser trained with a static oracle, \emph{gd} a greedy parser trained with a dynamic oracle, \emph{b} a beam search parser. Finally, the ``Strat'' column describes the strategy followed ($bu$=bottom-up, $td$=top-down and $in$=in-order). 
}
\label{tab:results}
\end{center}
\end{table}

\begin{table}
\begin{center}
\centering

\begin{tabular}{@{\hskip 0.5pt}lcccc@{\hskip 0.5pt}}
Parser & Bin & Type & Strat & F1 \\
\hline
\citet{Wang2015} & y & gs & bu & 83.2  \\
\citet{Liu2017}  & y & gs & bu & 85.7  \\
\textbf{This work} & n & gs &  bu & \textbf{86.3}\\
\textbf{This work} & n & gd &  bu & \textbf{86.8}\\
\hline
\citet{Zhu13} & y & b & bu & 83.2  \\
\citet{Watanabe15} & y & b & bu & 84.3  \\
\citet{Liu2017B} & y & b & bu & 85.5    \\
\hline
\citet{Dyer2016} & n & gs & td & 84.6    \\
\citet{Liu2017}  & n & gs & in & 86.1   \\
\hline
\multicolumn{1}{c}{}\\
\end{tabular}

\caption{Accuracy comparison of state-of-the-art shift-reduce constituent parsers on CTB~\S271-300. The ``Bin'' column marks if prior explicit binarization is required ($y$es/$n$o). The ``Type'' column shows the type of parser: \emph{gs} is a greedy parser trained with a static oracle, \emph{gd} a greedy parser trained with a dynamic oracle, and \emph{b} a beam search parser. Finally, the ``Strat'' column describes the strategy followed ($bu$=bottom-up, $td$=top-down and $in$=in-order). 
}
\label{tab:results2}
\end{center}
\end{table}

\begin{table}
\begin{center}
\centering
\begin{tabular}{lcc}
Parser & sent./s. & tran./sent. \\
\hline
Binary Bottom-up &  35.87 & 50.57 \\
Top-down &  38.78 & 59.76 \\
In-order &  33.34 & 60.83 \\
\textbf{This work} & \textbf{42.02} & \textbf{42.70}\\
\hline
\multicolumn{1}{c}{}\\
\end{tabular}
\caption{Comparison of parsing speed (sentences per second, sent./s.) and transition sequence length per sentence (tran./sent.) for the most common transition systems on WSJ~\S23, excluding time spent on un-binarization for the binary bottom-up parser. 
}
\label{tab:speed}
\end{center}
\end{table}

\subsection{Results}
Table~\ref{tab:results} and Table~\ref{tab:results2} compare
our system's accuracy to other state-of-the-art shift-reduce constituent parsers on the WSJ and CTB benchmarks.
Our non-binary bottom-up parser, regardless the kind of oracle used for training, improves over all other bottom-up systems with greedy decoding, as well as the top-down system by \citet{Dyer2016}, on both languages. 
Our approach is only outperformed slightly on the WSJ by the in-order parser by \citet{Liu2017} and is on par with
a
binary bottom-up parser with beam-search decoding and enhanced with lookahead features \citep{Liu2017B}. In the CTB, our system achieves the best accuracy. 

It is worth mentioning that the in-order and binary bottom-up parsers implemented by \citet{Liu2017} and the top-down system by \citet{Dyer2016} can be directly compared to our approach since all of them are implemented under the same framework and trained with the same hyper-parameters as \citep{Dyer2016}.

Regarding the effect of the dynamic oracle, our system benefits more on CTB (0.5 points) than on WSJ (0.2 points), but a wider study of error-exploration strategies might help to increase the final accuracy.

Table~\ref{tab:speed} reports parsing speeds and transition sequence length per sentence on WSJ~\S23 of four different transition systems implemented under the same framework by \citet{Dyer2016}.\footnote{\added{Please note that the framework developed by \citet{Dyer2016} was not optimized for speed and, therefore, reported speeds are just used for comparison purposes, but they should not be taken as the best speed  that a shift-reduce parser (implemented under the described neural network architecture) can potentially achieve.}} All of them are implemented on the same neural framework and use the same training settings as \citep{Dyer2016} and speeds are measured on a single core of an Intel i7-7700 CPU @3.60GHz. As expected, our  parser is the fastest by a considerable margin, since all other transition systems need a longer transition sequence to perform the same task. \added{Please note that times in decoding are dominated by neural network computations and a negligible  fraction of total running time is consumed by the transition-based algorithm. Therefore, we can easily improve shift-reduce parsing performance by reducing the number of transitions, as this is what determines the number of times that the neural model is required for scoring, even if we undertake this by using an algorithm with quadratic worst-case time complexity.}

\begin{table}
\begin{center}
\centering
\begin{tabular}{@{\hskip 0.5pt}lcccc@{\hskip 0.5pt}}
Parser & F1 & sent./s. \\
\hline
\textbf{This work} & 91.7 & \textbf{42.02} \\
\citet{SternAK17} & 91.8 &  22.79 \\
\citet{GaddySK18} & 92.1 & 19.48 \\
\citet{KleinK18} & \textbf{93.6} & 27.77 \\
\citet{KleinK18}+ELMO & \textbf{95.1} & 7.82 \\
\hline
\multicolumn{1}{c}{}\\
\end{tabular}
\caption{\added{Accuracy and speed comparison of our approach against current state-of-the-art chart-based constituent parsers on WSJ~\S{23}.}
}
\label{tab:resultsSOTA}
\end{center}
\end{table}

\added{In order to put our approach into context, we provide a comparison against four of the current state-of-the-art systems in performing constituent parsing. In particular, these parsers make use of global chart decoding to achieve the highest accuracies to date on WSJ in cubic running time, with a large cost in speed in comparison to shift-reduce systems. In table~\ref{tab:resultsSOTA}, we can see how our parser, without further speed optimization, is significantly faster than the best chart-based models under the same single core CPU conditions. This difference in speed, as well as the capability of incrementally processing the input from left to right, makes bottom-up shift-reduce parsers appealing for some downstream NLP applications, such as real-time machine translation systems, that require to produce a practically immediate output while the input is still coming.}

\begin{table*}[h]
\centering
\begin{tabular}{lccccc}
Parser & \#1 & \#2 & \#3 & \#4 & \#5 \\
\hline
Bin. bottom-up & 90.94 & 88.65 & 84.36 & 77.24 & 79.54 \\
Top-down & 90.98 & 88.76 & 85.01 & 76.63 & 77.35 \\
In-order & \textbf{91.36} & \textbf{89.21} &  \textbf{85.15} & 77.08 & 79.02\\
\textbf{This work} & 91.26 & 89.09 &  84.47 & \textbf{77.51} & \textbf{79.77} \\
\hline
\multicolumn{1}{c}{}\\
\end{tabular}
\caption{F-score on constituents with a number of children ranging from one to five on WSJ~\S23. 
}
\label{tab:analysis}
\end{table*}

\subsection{Structure Analysis}

We undertake a structure analysis to get insight into why our non-binary system is outperforming the binary version when the latter benefits from a prior binarization (which simplifies the initial problem and provides head information). In Table~\ref{tab:analysis} we present the F-score obtained by each transition system on creating constituents with a number of children ranging from one (unaries) to five. We use the variant of our transition system trained by a static oracle to carry out a fair comparison. From the results, we can state that the proposed non-binary shift-reduce parser improves over the binary one not only on more-than-binary branches, but also in unary and binary structures. 

It is also worth mentioning that both bottom-up systems perform better on building constituents with four and five children than the state-of-the-art in-order parser, while the top-down transition system achieves the worst results on these structures. This can be explained by the fact that constituents with a large number of children require a longer sequence of transitions to be built and, therefore, are more prone to suffer from error-propagation. Since the in-order and top-down parsers consume a greater amount of transitions than the bottom-up algorithms to produce the exactly same phrase structure, the former are being penalized in larger constituents. This also proves the advantage of the proposed non-binary approach over the binary algorithm on large constituents, since the first manages to reduce the transition sequences required by the second to build a certain tree and, therefore, alleviate the effect of error-propagation.

From this simple experiment, we can also see that a binary bottom-up parser tends to be less accurate on unary and binary branches in comparison to a purely top-down strategy, while the latter suffers a drop in F-score when it comes to build constituents with four or more children. This also explains that, by combining both strategies as the in-order algorithm by \citet{Liu2017} does, we can build a more accurate parser. It also seems that the non-binary bottom-up transition system alleviates the weaknesses of the binary version on building constituents with a short number of children, while keeping a good accuracy on larger structures.

\subsection{\added{Empirical Running Time Complexity}}
\begin{figure}
\begin{center}
\small
\includegraphics[width=0.44\textwidth]{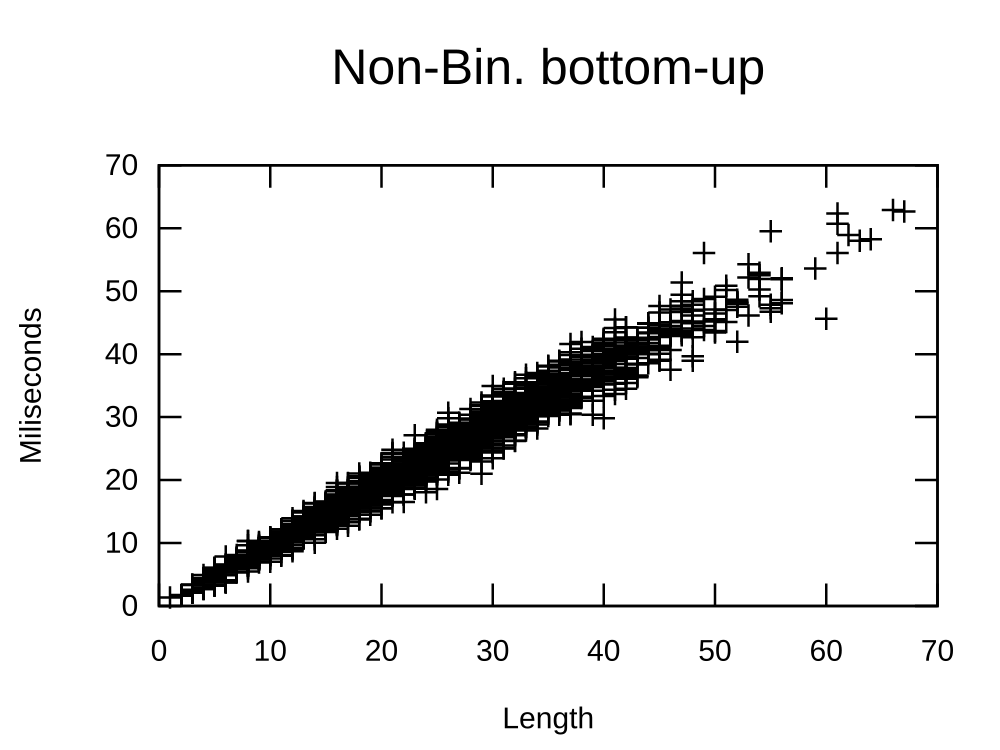}
\qquad\qquad
\includegraphics[width=0.44\textwidth]{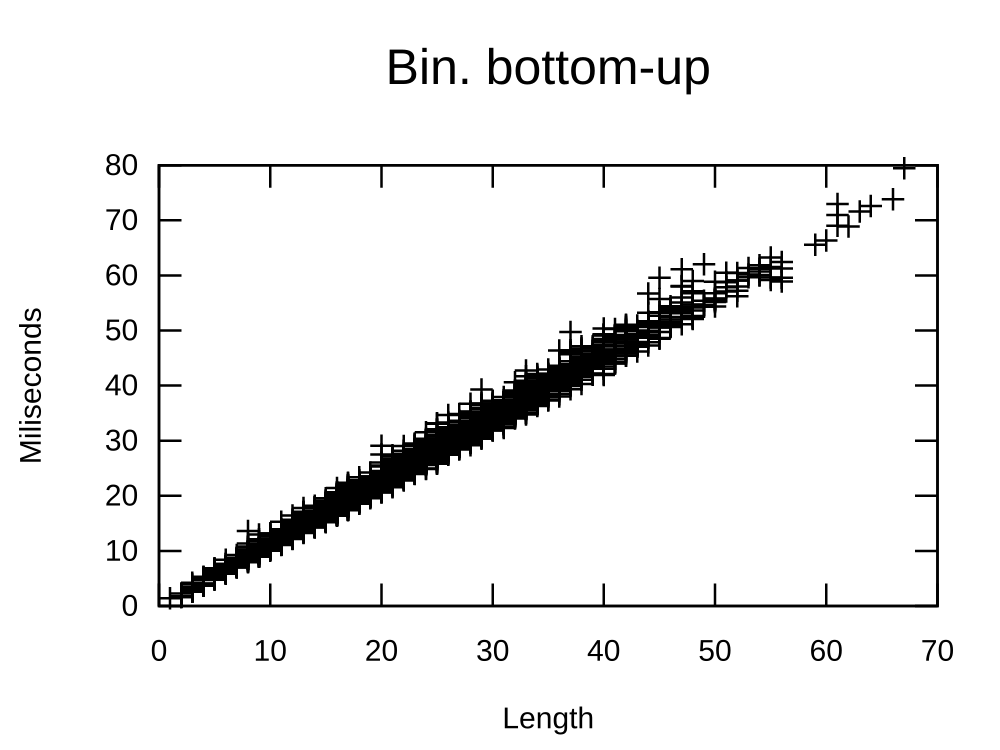}
\includegraphics[width=0.44\textwidth]{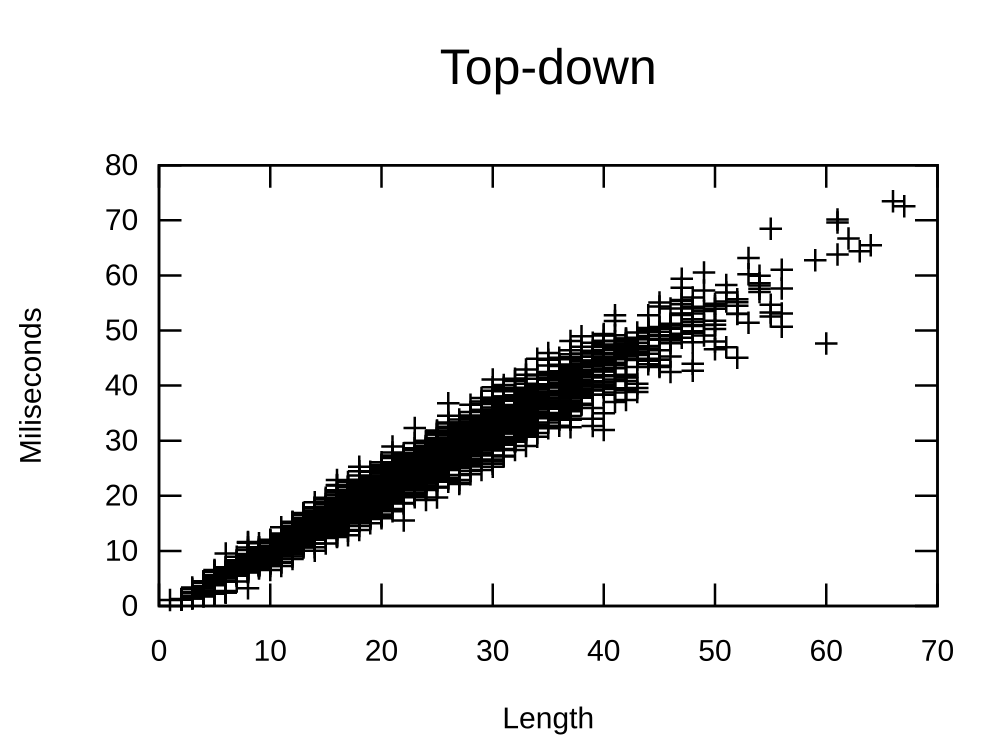}
\qquad\qquad
\includegraphics[width=0.44\textwidth]{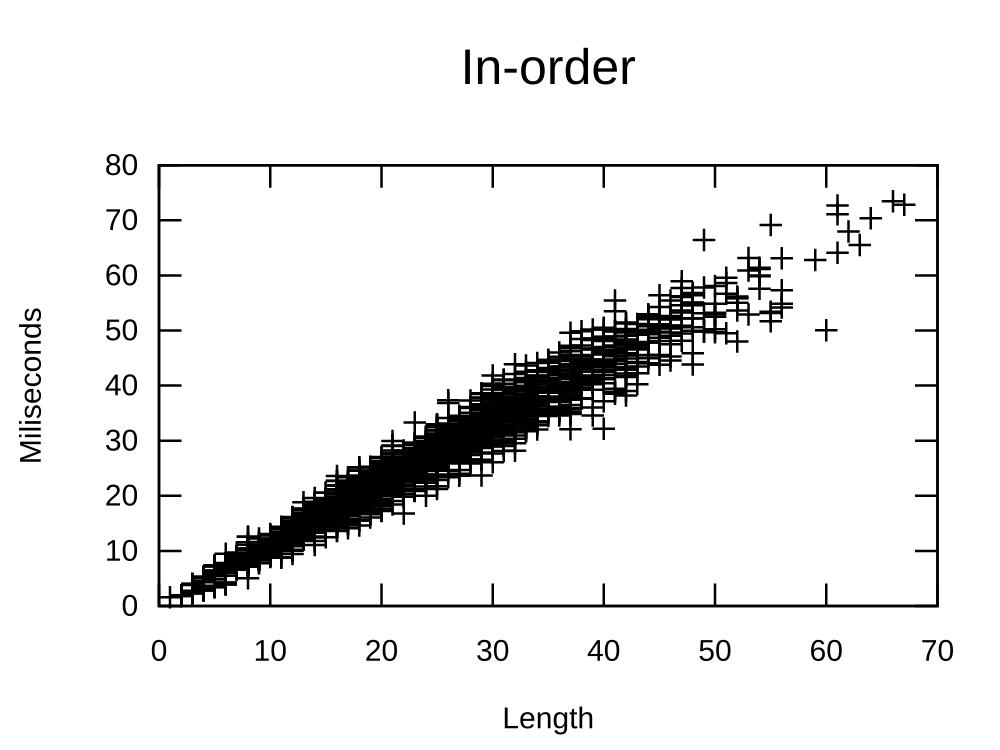}
\end{center}
\caption{\added{Running time relative to the length of the sentence, for the three mentioned lineal transition-based algorithms plus the novel non-binary shift-reduce parser on the 2,416 sentences from WSJ~\S23.}}
\label{fig:run}
\end{figure}

\added{We now show that, in spite of being a worst-case quadratic running time algorithm in theory, our approach performs as a linear parser during decoding in practice. In particular, we measure the time spent by the system to analyze every sentence from WSJ~\S23 and depict the relation between time consumed and sentence length. We repeat this experiment for the other three transition-based algorithms implemented under the same framework. As shown in Figure~\ref{fig:run}, a linear behaviour is observed in the non-binary parser, similarly to the other three linear algorithms, proving that the number of parametrized $\re$ transitions, evaluated by the model at each step, is considerably low (behaving practically like a constant). }
\label{sec:run}

\section{Conclusion}
\label{conclusion}
We present,
to our knowledge, 
the first
purely non-binary
bottom-up shift-reduce constituent parser and we also develop an optimal dynamic oracle for training it. Unlike traditional bottom-up systems, this novel algorithm can be applied to any language without the need of further additional resources, required to perform prior binarization, \added{and, despite being quadratic in theory, it performs as a linear-time parser in practice}.

Except the in-order parser by \citet{Liu2017} on the WSJ, it outperforms all other greedy shift-reduce parsers in terms of accuracy with just static training, and matches the second best result on the WSJ when we use a dynamic oracle for training, on par with the system developed by \citet{Liu2017B}, which uses beam search and is enhanced with lookahead features. In addition, our system obtains the highest accuracy on CTB, regardless of the oracle used for training. 

Additionally, we note that our algorithm is the fastest transition system developed so far for constituent parsing, as it consumes the shortest sequence of transitions to produce phrase-structure trees. In practice, it outspeeds other approaches in a comparison under homogeneous conditions and it will certainly alleviate the bottleneck caused by parsers in NLP applications that rely on syntactic representations.

Finally, we also prove that the novel non-binary algorithm excels in building trees with a large number of children. This is probably due to the fact that our approach requires a shorter number of transitions to build a constituent and, therefore, unlike the other existing transition systems, it is less prone to suffer from the error propagation generated when a long sequence of actions are applied to build these kind of structures.

The parser's source code will be freely available after acceptance.

\section*{Acknowledgments}
This work has received funding from the European
Research Council (ERC), under the European
Union's Horizon 2020 research and innovation
programme (FASTPARSE, grant agreement No
714150), from the ANSWER-ASAP project (TIN2017-85160-C2-1-R) from MINECO, and from Xunta de Galicia (ED431B 2017/01).

\section*{References}

\bibliography{main,twoplanaracl,bibliography}

\end{document}